\documentclass[runningheads]{llncs}
\usepackage{graphicx}
\usepackage{amsmath,amssymb} 

\usepackage[utf8]{inputenc} 
\usepackage[T1]{fontenc}    
\usepackage{hyperref}       
\usepackage{url}            
\usepackage{booktabs}       
\usepackage{amsfonts}       
\usepackage{nicefrac}       
\usepackage{microtype}      

\usepackage{lmodern}
\usepackage{enumerate}
\usepackage[usenames]{color}
\usepackage{capt-of}
\usepackage{epstopdf}
\graphicspath{{fig_paper/}}
\usepackage{wrapfig}
\usepackage[font=scriptsize,labelfont=bf]{caption}
\usepackage{multirow}
\usepackage[linesnumbered,ruled]{algorithm2e}
\usepackage{algorithmic}

\usepackage{subfig}
\usepackage{tabularx}
\usepackage{afterpage}
\usepackage{floatflt}
\usepackage{array}
\usepackage{arydshln}

\newcommand*\patchAmsMathEnvironmentForLineno[1]{%
	\expandafter\let\csname old#1\expandafter\endcsname\csname #1\endcsname
	\expandafter\let\csname oldend#1\expandafter\endcsname\csname end#1\endcsname
	\renewenvironment{#1}%
	{\linenomath\csname old#1\endcsname}%
	{\csname oldend#1\endcsname\endlinenomath}}%

\newcommand{\inner}[1]{\left\langle#1\right\rangle}

\def\R{\mathbb{R}}

\def\S{\mathcal{S}}

\newcommand{\norm}[1]{\left\|#1\right\|}

\def\argmax{\mathop{\rm arg\,max}\limits}
\def\argmin{\mathop{\rm arg\,min}\limits}
\def\minop{\mathop{\rm min}\limits}
\def\maxop{\mathop{\rm max}\limits}
\def\sign{\mathop{\rm sign}\limits}

\def\max{\mathop{\rm max}\nolimits}

\newcommand\ChangeRT[1]{\noalign{\hrule height #1}}

\begin{document}

\title{A randomized gradient-free attack on ReLU networks}

\titlerunning{A randomized gradient-free attack on ReLU networks}
%
\author{Francesco Croce\inst{1} \and
Matthias Hein \inst{2}}

\authorrunning{F. Croce and M. Hein}

\institute{Department of Mathematics and Computer Science,\\ Saarland University, Germany
	\and Department of Computer Science,\\ University of T\"ubingen, Germany}

\maketitle              

\begin{abstract}
	It has recently been shown that neural networks but also other classifiers are vulnerable to so called adversarial attacks e.g. in object recognition an almost non-perceivable change of the image changes the decision of the classifier. Relatively fast heuristics have been proposed to produce these
	adversarial inputs but the problem of finding the optimal adversarial input, that is with the minimal change of the input, is NP-hard.  While methods based on mixed-integer optimization which find the optimal adversarial input have been developed, they do not scale to large networks. Currently, the attack scheme
	proposed by Carlini and Wagner is considered to produce the best adversarial inputs. In this paper we propose a new attack scheme for
	the class of ReLU networks based on a direct optimization on the resulting linear regions. In our experimental validation we improve in all except one
	experiment out of 18 over the Carlini-Wagner attack with a relative improvement of up to 9\%.  As our approach is based on the geometrical structure of 
	ReLU networks, it is less susceptible to defences targeting their functional properties.
	\keywords{adversarial manipulation, robustness of classifiers.}
\end{abstract}

\section{Introduction}
In recent years it has been highlighted that state-of-the-art neural networks are highly non-robust: small changes to an input image, which are basically non-perceivable for humans, change the classifier decision and the wrong decision has even high confidence \cite{SzeEtAl2014,GooShlSze2015}. This calls into question the usage of neural networks in safety-critical systems e.g. medical diagnosis systems or self-driving cars and opens up possibilities to actively attack an ML system in an adversarial way \cite{PapEtAl2016a,LiuEtAl2016a,KurGooBen2016a}. Moreover, this non-robustness has also implications on follow-up processes like interpretability. How should we be able to interpret classifier decisions if very small changes of the input lead to different decisions?

This started an arms race looking for both effective defences against adversarial manipulations and new, sophisticated attacking methods to generate 
adversarial inputs \cite{YuanEtAL2017}. In the end it turned out that most proposed defences were proven to be ineffective against attacks of type other than that considered by the defence itself \cite{CarWag2017,AthEtAl2018}. Regarding adversarial attacks one can distinguish between exact methods which find the optimal adversarial
input with respect to different $l_p$-norms e.g. wrt to  $l_1$-norm \cite{CarEtAl2017}, $l_\infty$-norm \cite{KatzEtAl2017} and recently $l_p$ for $p=1,2,\infty$ in \cite{TjeTed2017}. The problem is in general NP-hard \cite{KatzEtAl2017} and can be formulated as a mixed-integer optimization problem \cite{TjeTed2017} 
which does not scale to large networks. Therefore early on fast heuristics were proposed to produce adversarial attacks \cite{GooShlSze2015,HuaEtAl2016,MooFawFro2016,MadEtAl2018} as this is important when one wants to use them in \emph{adversarial training} to robustify the networks. More recently, there has been a focus on methods in the middle range which are close the optimal adversarial input but are still scalable to 
larger networks, where currently the attack of Carlini and Wagner (CW-attack) \cite{CarWag2016} is considered to be state-of-the-art.

Another line of research provides robustness guarantees by giving lower bounds on the norm of the minimal perturbation necessary to change the class
\cite{HeiAnd2017,RagSteLia2018,WonKol2018} but these methods are restricted to one hidden layer \cite{HeiAnd2017,RagSteLia2018}
or rather shallow networks \cite{WonKol2018} and do not provide adversarial examples.

In this paper we propose a new method for the generation of adversarial inputs for ReLU feedforward neural networks which exploits only the underlying geometric structure of the classifier which are well known to result in continuous piecewise affine functions  \cite{AroEtAl2018}. We show that on each region where the classifier is affine (linear region) the problem of finding adversarial inputs is thus a quadratic program for $p=2$. However, as the number of linear regions grows exponentially with depth it is typically impossible to check all of them and thus we propose an efficient exploration strategy.  Finally, because of its geometric nature and our random exploration scheme our attack is more resistant to defences which are based on functional properties of the model like gradient based ones.
It produces adversarial inputs of high quality outperforming the current state-of-the-art  CW-attack \cite{CarWag2016} on almost all architectures and three data sets which we tested with a relative improvement of up to 9\%.

\section{Local linear representation of ReLU networks}\label{sec:linear}
In the following we consider multi-class classifiers $f:\R^d \rightarrow \R^K$ where $d$ is the input dimension and $K$ the number of classes.
It is well known that feedforward neural networks which use ReLU activation functions and are linear in the output layer yield continuous piecewise affine functions for each output component, see e.g. \cite{AroEtAl2018}. Also residual and convolutional networks with ReLU (as well as leaky ReLU) activation function or max/sum/avg pooling layers lead to continuous piecewise
affine classifier functions. However, as these architectures require special care, for simplicity we restrict ourselves to fully connected neural networks.
\begin{definition} A function $h:\R^d \rightarrow \R$ is called \emph{piecewise affine} if there exists a finite set of polytopes $\{Q_r\}_{r=1}^M$ 
	(linear regions of $h$) such that $\cup_{r=1}^M Q_r= \R^d$ and $h$ is an affine function when restricted to  every $Q_r$.
\end{definition}
\noindent
We derive now, given an input point $x$, the explicit formulation of the linear counterpart of $f$ for $x$ and the polytope $Q(x)$ to which $x$ belongs (and on which the $f$ is an affine function). Note that this 
description need not be unique when $x$ lies on the boundary between two or more linear regions. But as this is a set of measure zero it does not play any role in practice.\\
Let $L+1$ be the total number of layers and denote the weights and biases by $W^{(l)} \in\mathbb{R}^{n_l \times n_{l-1}}$ and $b^{(l)} \in \R^{n_l}$  for $l=1,\ldots,L+1$ where $n_0=d$ and $n_{L+1}=K$.
We define the feedforward network in the usual recursive way, where for $x\in \R^d$ we define $g^{(0)}(x)=x$ and the output of the $k$-th layer before and after applying componentwise the ReLU activation function $\sigma:\R \rightarrow \R$, $\sigma(t)=\max\{0,t\}$ as
\[f^{(k)}(x)=W^{(k)}g^{(k-1)}(x)+b^k, \quad \mathrm{ and } \quad g^{(k)}(x)=\sigma(f^{(k)}(x)), \quad k=1,\ldots,L.\]
The final classifier function is given as $f^{(L+1)}(x)=W^{(L+1)}g^{(L)}(x) + b^{(L+1)}$ and decisions are made using $\argmax_{r=1,\ldots,K} f^{(L+1)}_r(x)$.
We define the diagonal matrices $\Delta^{(l)},\Sigma^{(l)}\in \R^{n_l \times n_l}$ for $l=1,\ldots,L$ with entries defined as 
\begin{gather*} \Delta^{(l)}(x)_{ij} = \begin{cases} \sign(f_i^{(l)}(x)) & \textrm{if } i=j,\\ 0 & \textrm{else.} \end{cases}, \hspace{5pt}
\Sigma^{(l)}(x)_{ij} = \begin{cases} 1 & \textrm{if } i=j, f_i^{(l)}(x)>0,\\ 0 & \textrm{else.} \end{cases}.\end{gather*}
The matrices $\Delta^{(l)}$ are used later to derive the boundaries of the polytope, while the matrices $\Sigma^{(l)}$ are the linear equivalent to ReLU functions at a specific point.
This allows us to write $f^{(k)}(x)$ in terms of matrix and vector products as
\[ f^{(k)}(x)=W^{(k)}\Sigma^{(k-1)}\Big(W^{(k-1)} \Sigma^{(k-2)} \Big(\ldots  \Big( W^{(1)}x + b^{(1)}\Big) \ldots\Big)+ b^{(k-1)} \Big) + b^{(k)}.\]
This can be further rewritten as $f^{(k)}(x) = V^{(k)}x + a^{(k)}$, with $V^{(k)} \in \R^{n_k \times d}$ and $a^{(k)} \in \R^{n_k}$, where
\begin{gather*} V^{(k)} = W^{(k)} \prod_{l=1}^{k-1} \Sigma^{(k-l)}W^{(k-l)},\quad a^{(k)} = b^{(k)} + \sum_{l=1}^{k-1} \Big(\prod_{m=1}^{k-l} W^{(k+1-m)} \Sigma^{(k-m)}\Big) b^{(l)}. \end{gather*}
Note that $V^{(k)}$ and $a^{(k)}$ are accessible via a forward-pass through the network with only slightly increased overhead compared to the normal effort required for computing the output for a given input $x$.
The polytope $Q(x)$ can now be described by the inequalities
\[ Q(x) = \big\{z \in \R^d \,\Big|\, \Delta^{(l)}(x)\big(V^{(l)}z +a^{(l)}\big)\geq 0, \quad l=1,\ldots,L \big\},\]
where the diagonal matrix $\Delta^{(l)}(x)$  contains the information on which part of the hyperplane $x$ lies. 
In total one gets $N=\sum_{l=1}^L n_l$ inequality constraints
describing the polytope and $N$ represents also the total number of hidden neurons of the network. We denote the set of the inequalities defining $Q(x)$ as $\S(Q(x))$.

Finally, the multi-class classifier is described on $Q(x)$ by the affine function 
\[ f^{(L+1)}=V^{(L+1)}x + a^{(L+1)}. \label{eq:lin_classifier}\]
Then the decision region for a class $c$ on $Q(x)$ is the polytope $P_c$ described by $K-1$ inequalities
\[ P_c=\big\{z \in \R^d \,\big|\,\inner{V^{(L+1)}_c -V^{(L+1)}_s, z } + a^{(L+1)}_c - a^{(L+1)}_s \geq 0 \quad \forall s \neq c\big\} \cap Q(x),\]
where $V^{(L+1)}_r$ is the $r$-th row of $V^{(L+1)}$. The set $P_c$ is again a polytope since obtained as intersection of two polytopes. Note that the intersection with  $Q(x)$ is necessary as the linear description for the classifier obtained for $x$ is only valid on $Q(x)$.\\
In dependency of the application domain of the classifier there might be also other constraints which the input has to fulfill such as box constraints in the case of images. In this case, in all the definitions one has to take the intersection with this constraint set. Without much loss of generality we assume that such a
set $M$ is also given as a polytope so that the intersections $Q(x) \cap M$ and $P_c \cap M$ are polytopes as well.

\section{Generation of adversarial inputs on the linear regions}\label{sec:adv}
We first define the optimization problem for generating adversarial inputs, that is given a multi-class classifier $f:\R^d \rightarrow \R^K$ and an input $x$
we want to find the smallest perturbation $x+\delta$ such that the decision of $f$ for $x+\delta$ is different from that of $x$. Typical $l_p$ metrics with
$p \in \{1,2,\infty\}$ are used to measure the ``size'' of the perturbation. In the experiments we choose $p=2$, but the framework below can be easily extended to
$p \in \{1,\infty\}$.

Formally,  the problem can be described as the following optimization problem \cite{SzeEtAl2014}. 
Suppose that the classifier outputs class $c$ for input $x$, that is $f_c(x)=\argmax_{j=1,\ldots,K} f_j(x)$
(we assume the decision is unique).  The problem of generating the minimal perturbation $x+\delta$ 
such that the classifier decision changes is equivalent to finding the solution of
\begin{align}\label{eq:advopt}
\minop_{\delta \in \mathbb{R}^d} \; \norm{\delta}_p, \qquad \textrm{s.th.}  \quad    \maxop_{l\neq c} \; f_l(x+\delta) \geq f_c(x+\delta) \; \textrm{ and }\; x+\delta \in C,
\end{align}
where $C$ is a constraint set which the generated point $x+\delta$ has to satisfy, e.g., an image has to be in $[0,1]^d$ but more generally we assume
$C$ to be a polytope. 
The complexity of the optimization problem \eqref{eq:advopt} depends on the classifier $f$, but it is typically non-convex (see \cite{KatzEtAl2017} for a hardness result for ReLU networks).
Many relaxations and approximate solutions of problem \eqref{eq:advopt} have been proposed in the literature in order to make it computationally tractable, see e.g. \cite{MooEtAl2016,CarWag2016}, and for ReLU networks a few formulations have been proposed for the exact solution \cite{KatzEtAl2017,CarEtAl2017,TjeTed2017}, most notably \cite{TjeTed2017} which uses mixed-integer programming to achieve this. 

However, as we show in the following, the optimization problem \eqref{eq:advopt} can be solved efficiently on any linear region of $f$. 
In fact, given a point $y$ in the input space we recall from Section \ref{sec:linear} that in the polytope $Q(y)$, or equivalently inside the linear region containing $y$ and where the classifier is affine, it holds $f(z)=Vz + a$ for some $V,a$. Thus the optimization problem
\begin{align}\delta=\argmin_{l\neq c}\norm{\delta_l}_p, \label{eq:advopt_multiclass} \end{align} with $\delta_l$ being the solution of
\begin{align}\label{eq:advopt_lin}
\begin{split} \minop_{\delta \in \mathbb{R}^d} \; \norm{\delta}_p, \quad \textrm{s.th.} \hspace{5pt}    & \inner{V_l-V_c,x+\delta} + a_l-a_c \geq 0,\hspace{5pt} x+\delta \in C \cap Q(y), \end{split}
\end{align}
where $V_i$ and $a_i$ are respectively the $i$-th row and component of $V$ and $a$, is equivalent to \eqref{eq:advopt} on $Q(y)$. 
First of all we note that when the class $l$ is fixed and $p=2$ the resulting optimization problem on $Q(y)$ is a convex quadratic program, whereas for $p \in \{1,\infty\}$ it is a linear
program. Further constraints in form of a polytope e.g. box constraints on the input space do not change the type of the optimization problem.
Note that it can in principle happen that there exists no feasible point if the decision boundary of $f$ does not intersect the linear region $Q(y)$. Nonetheless, typically one gets at least for one class an adversarial input in this way.

Although the number of linear regions grows exponentially with the size of the networks \cite{MonEtAl2014,AroEtAl2018}, it is finite. Thus one can in principle solve \eqref{eq:advopt} by checking all the linear regions. However, this is clearly
infeasible for all possible regions. Thus we develop in the following a mixed randomized strategy of exploration and local search which solves
\eqref{eq:advopt_multiclass} just on a small number of linear regions but nevertheless achieves good quality.

\section{A linear regions based method to generate adversarial examples}\label{sec:our_method}
\begin{algorithm}
	\SetAlgoNoLine
	\caption{rLR-QP}
	\label{alg:algorithm-label}
	\SetKwInOut{Input}{Input}
	\SetKwInOut{Output}{Output}
	\Input{$x$ original image, $\delta_{WS}$ starting perturbation, $L$ set of target classes,\\ $n_1,n_2,n_3,n_4,\alpha$ hyperparameters}
	\Output{$\delta$ adversarial perturbation}
	$\delta_0 \gets$ solution of problem \eqref{eq:advopt_multiclass} on $Q(x)$ and $l\in L$\\
	\leIf{$\norm{\delta_0}_2<\norm{\delta_{WS}}_2$}{$\delta \gets \delta_0$, $u\gets \norm{\delta_0}_2$}{$\delta \gets \delta_{WS}$, $u\gets \norm{\delta_{WS}}_2$}
	Initialize the set $S$ of $n_1$ copies of $\delta$\\
	\For{$a_1=1,...,n_4$}{
		\underline{\textbf{Exploration Step:}}\\
		\ForEach{$a_2=1,...,n_3$}{
			$R=\emptyset$\\
			\For{$i=1,...,n_1$}{\ForEach{$j=1,...,n_2$}{$\epsilon\gets$ random point in $B(0,u/a_1)$\\ $R\gets R \cup \{x + S_i + \epsilon\}$}}
			\ForEach{$y\in R$}{$\delta_{temp}\gets$ solution of problem \eqref{eq:advopt_multiclass} on $Q(y)$ and $l\in L$\\ \lIf{$\norm{\delta_{temp}}_2<u$}{$\delta \gets \delta_0$, $u\gets\norm{\delta}_2$}
				$m=\argmax_{i=1,...,n_1}\norm{S_i}_2$\\
				\lIf{$\norm{\delta_{temp}}_2<\alpha u$ \emph{and} $\norm{\delta_{temp}}_2<\norm{S_m}_2$}{Replace $S_m$ with $\delta_{temp}$}}}
		\underline{\textbf{Local Search Step:}}\\
		\ForEach{$j=1,...,n_2$}{$\epsilon\gets$ random point in $B(0,u/a_1)$\\ $y\gets x + \delta + \epsilon$\\ $\delta_{temp}\gets$ solution of problem \eqref{eq:advopt_multiclass} on $Q(y)$ and $l\in L$\\ \lIf{$\norm{\delta_{temp}}_2<u$}{$\delta \gets \delta_0$, $u\gets\norm{\delta}_2$}
			$m=\argmax_{i=1,...,n_1}\norm{S_i}_2$\\
			\lIf{$\norm{\delta_{temp}}_2<\alpha u$ \emph{and} $\norm{\delta_{temp}}_2<\norm{S_m}_2$}{Replace $S_m$ with $\delta_{temp}$}}}
\end{algorithm}

As pointed out in the previous section, solving a finite number of problems like \eqref{eq:advopt_lin} suffices to solve problem \eqref{eq:advopt}, but enumerating and checking all the linear regions of a generic classifier is not feasible. In the following we motivate our randomized strategy together with several ways to speed-up the computations by either avoiding to solve the QPs by simple checks or how to speed up the QP computation itself. 

It is clear that the solution of the QP \eqref{eq:advopt_lin} is typically attained at a face of $Q(y)$ and the solution lies on the decision boundary. At first sight it
looks as the optimal next step, after finding a first adversarial input, is to just visit the neighboring polytope to track the decision boundary. Unfortunately, this approach does not work in practice
as the solution is attained typically at a face of dimension $m<d-1$ (more than one linear
constraint of the polytope is active). Thus the neighboring region is not uniquely defined and the number of neighboring regions grows roughly exponentially in $d-1-m$. Then, one would have to check all of them and this is again infeasible. But even if this was tractable, nevertheless one would still need a strategy in case all of these
neighboring regions just lead to worse adversarial inputs. 

This motivates our randomized strategy to select linear regions which is based on an exploration step, in order to get out of a potential suboptimal valley, and a local search strategy, where we just check linear regions in the close vicinity of the current best adversarial input. Please note that we select linear regions by generating randomly  points $y$ and then check the corresponding linear region $Q(y)$. Note also that in this way we will be biased to visiting linear regions of high volume, which makes sense as in such way the portion of the input space checked with a limited number of regions tends to be maximized. 

The final scheme is summarized in Algorithm \ref{alg:algorithm-label}, which we refer to as \emph{rLR-QP}. Please note that $\delta$ is the best adversarial input found so far and $u=\norm{\delta}_2$. Another norm than
$p=2$ can be simply optimized by changing the corresponding convex optimization problem in \eqref{eq:advopt_lin}. We have the possibility to use adversarial inputs from another method as initialization. Later on we initialize our method with \textit{DeepFool} \cite{MooEtAl2016} to speed up the procedure. As a first step the algorithm checks the linear region $Q(x)$ 
which contains the original input $x$. This region contains in some cases already the optimal adversarial input. After this initialization phase the algorithm alternates between the exploration step and the local search step. In the exploration step we keep a set of adversarial inputs which are up to $\alpha$ times suboptimal, where $\alpha=1.5$ in all experiments, and search in their neighborhood. This exploration phase prevents the method to be too much focused on only one region and explores a larger neighborhood of $x$. The second phase is the local search step where we search in a smaller neighborhood around the best adversarial input found so far. Note that our random sampling on $B(0,u)$ is uniform in the direction
and uniform in the radius (this means we are \emph{not} sampling from the uniform distribution on $B(0,u)$). The reason is that we want to have a higher probability for checking also
smaller radii, whereas for uniform sampling in high dimensions the samples would be very close to the surface of the sphere.

The parameters $n_1,n_2,n_3,n_4$ can be used to tune the total number of linear regions that are explored and consequently the runtime of the algorithm. In fact, we have that overall $M=1+(n_1\cdot n_3 + 1)\cdot n_2\cdot n_4$ points are picked during the process, so that $M$ is an upper bound on how many linear regions are checked (we make sure not to check the same region twice). For the experiments we fix $n_1,n_2=10$, $n_3=5$, $n_4=3$.


\section{Experiments}
In the experiments we compare the performances of our method \emph{rLR-QP} with different state-of-the-art methods. First, we compare our algorithm with an exact method for a small networks guaranteed to find the optimal adversarial input. Second, we apply our method on larger networks, trained both on the original training set and with adversarial training as described in \cite{MadEtAl2018}. Finally, we test how the number of points checked by \emph{rLR-QP} improves the quality of the outcome. In all cases, we analyse our results in relation to those obtained by the Carlini-Wagner $l_2$-attack \cite{CarWag2016} (\textit{CW}), as it is considered to be state-of-the-art method to produce high quality adversarial examples, and \textit{DeepFool} \cite{MooEtAl2016}, as it is a very fast method providing already good to high quality adversarial examples which we use as initialization.

\subsubsection{Comparison with provably optimal solutions}
\begin{figure}[t]
	\centering
	\includegraphics[scale=0.45]{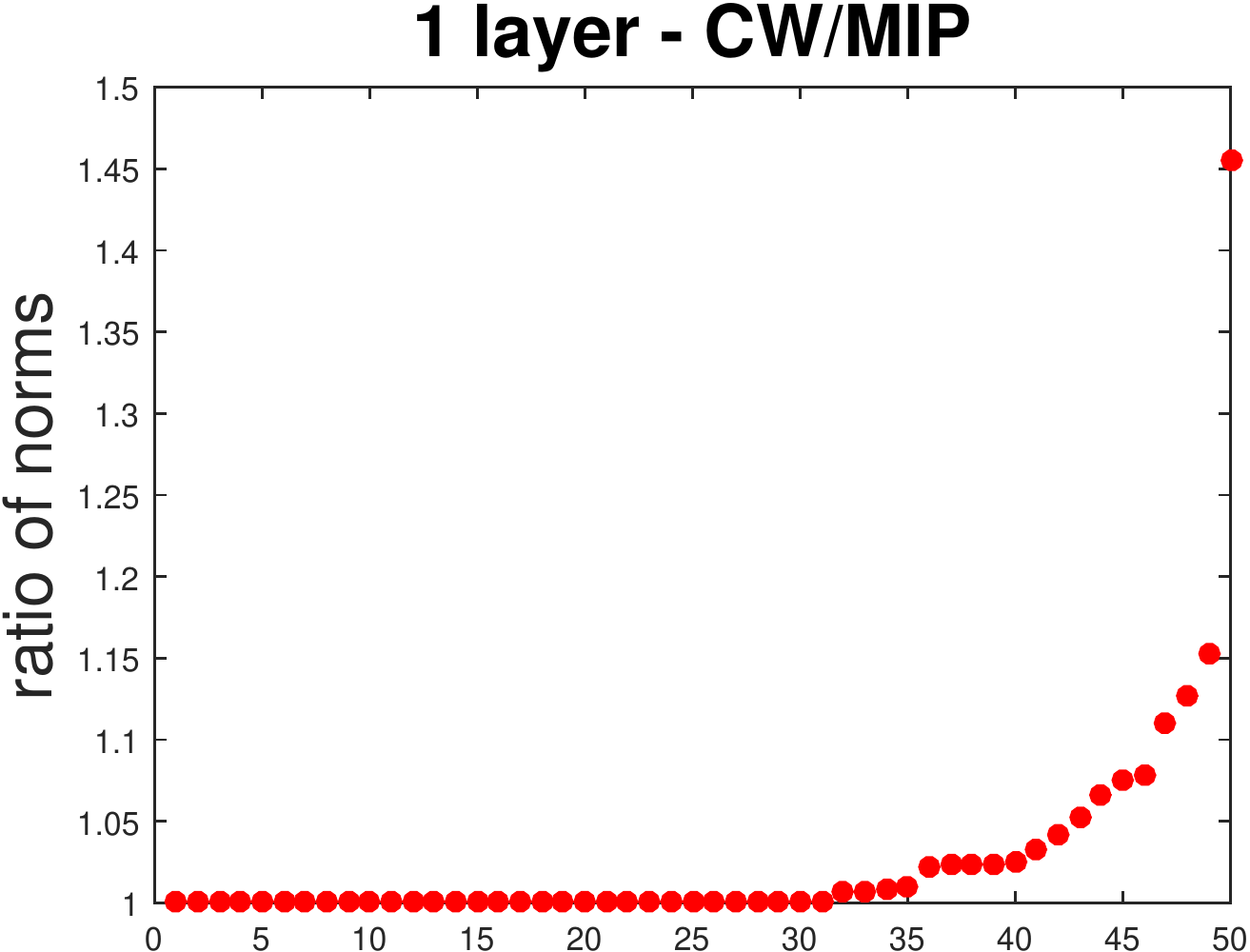}
	\includegraphics[scale=0.45]{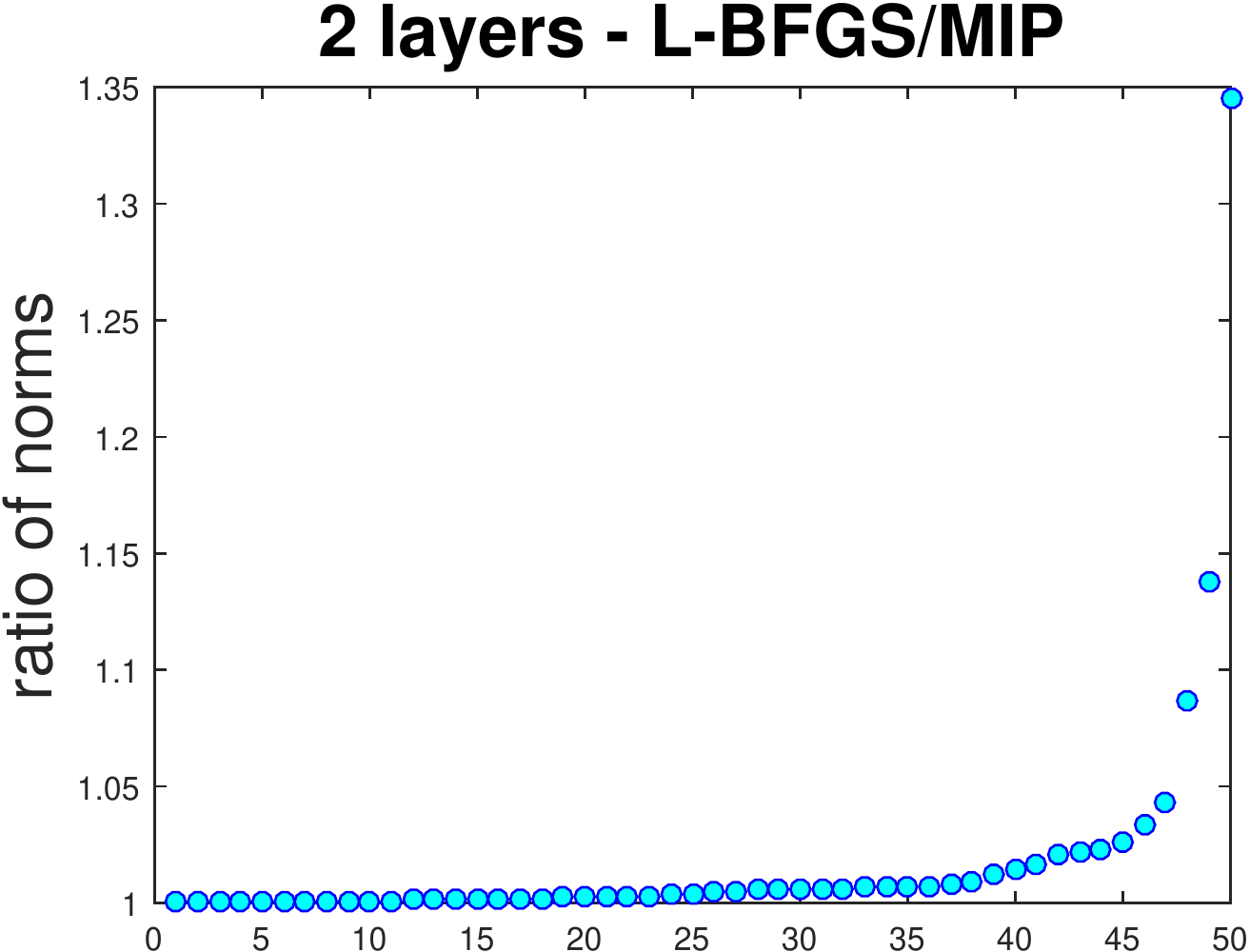}
	\caption{\textbf{Effectiveness of attacks.} Sorted ratios of the norm of the perturbation found by the attacking methods over the norm of the optimal solution provided by \textit{MIP} (computed on the first 50 images of MNIST test set). \textbf{Left:} performance of \textit{CW} $l_2$-attack on network $N_1$. \textbf{Right:} performance of \textit{L-BFGS} on network $N_2$. In both cases one observes that on about 70\% of the points the attacks find solutions close to the optimal ones, while there are approximately 30\% of examples where the difference becomes significant and for a few of them even large.}
	\label{fig:distr_MIP}
\end{figure}

\begin{table}[t]
	\newcolumntype{C}[1]{>{\centering\arraybackslash}p{#1}}
	\centering
	\caption{\textbf{Relative quality of generated adversarial examples.} We report the average and the maximum value of the ratios between the 2-norms of the adversarial inputs found by the indicated method and the optimal adversarial inputs generated by the \textit{MIP} method \cite{TjeTed2017} (smaller is better). Two neural networks $N_1$ (one hidden layer), $N_2$ (two hidden layers) have been trained on MNIST. The statistics have been computed with the first 50 images of the test set of MNIST. Please notice that our method finds the optimal solution for all 50 test images. All other methods are, sometimes quite significantly, worse.\\}
	\begin{tabular}{C{8mm}|C{12mm} C{12mm}|C{12mm} C{12mm}|C{12mm} C{12mm}|C{12mm} C{12mm}}
		& \multicolumn{2}{c}{\emph{rLR-QP} (ours)} & \multicolumn{2}{c}{\textit{DeepFool}}& \multicolumn{2}{c}{\textit{L-BFGS}}& \multicolumn{2}{c}{\textit{CW}}\\
		\hline\hline 
		& mean & max & mean & max & mean & max & mean & max \\
		\hline
		$N_1$& \textbf{1.0000} & \textbf{1.0000} & 1.1226 & 1.6059 & 1.2749 & 2.2936 & 1.0269 & 1.4546 \\
		$N_2$& \textbf{1.0000} & \textbf{1.0000} & 1.1426 & 1.5953 & 1.0180 & 1.3457 & 1.0686 & 1.4517
	\end{tabular}
	\label{tab:MIP}
\end{table}

In \cite{TjeTed2017} problem \eqref{eq:advopt} is solved using a mixed integer programming reformulation. Since we want to evaluate the quality of adversarial examples found by our method (\emph{rLR-QP}), we compare them to the minimal solutions obtained as in \cite{TjeTed2017} (\textit{MIP}). Moreover, we also report the performances of other three attacks: \textit{DeepFool}, \textit{L-BFGS} \cite{SzeEtAl2014} and the \textit{CW} $l_2$-attack.
We train two small neural networks on MNIST dataset: $N_1$ with 1 hidden layer of 50 units and $N_2$ with 2 hidden layers of 50 and 15 units. For our method we use the outcome of \textit{DeepFool} as starting point and perform the search over all the 9 target classes. We consider the first 50 images of the test set for the evaluation of adversarial inputs.

We report both the average and the maximum of the ratios \nicefrac{$\norm{\delta}_2$}{$\norm{\delta_{MIP}}_2$}, where $\delta$ is the perturbation provided by a certain attack while $\delta_{MIP}$ is the provably optimal solution of \eqref{eq:advopt} given by \textit{MIP}. As Table \ref{tab:MIP} shows, \emph{rLR-QP} finds the optimal solution for all 50 test inputs. The other methods instead perform worse and even when the mean is not far from 1 the maximum reaches high values. This behavior is illustrated in Figure \ref{fig:distr_MIP} where all the 50 ratios are reported for both \textit{CW} on $N_1$ and \textit{L-BFGS} on $N_2$ in increasing order. These are the two settings with the \emph{best} mean performance ($N_1$/\textit{CW}: 1.027, $N_2$/\textit{L-BFGS}: 1.018). We see in Figure \ref{fig:distr_MIP} that in both cases there are a few outliers on which the performance of the attacks is significantly
worse compared to the optimum.

Since it is difficult to provide a fair, complete comparison of the runtime of different methods, we highlight that \emph{rLR-QP} takes on average around 2 seconds on $N_1$ and less than 0.5 seconds on $N_2$ to perform on a single image, without any sort of parallelization (see below for further details). We have a shorter time on a more complex network for different reasons. First, the time needed to solve the QPs depends significantly to the size of the optimal solutions themselves (smaller perturbations are usually faster to be computed).  Moreover, we did not conduct a wide search in order to tune the hyperparameters $n_1,...,n_4$ described in Section \ref{sec:our_method}, so one might get the same results with smaller computational effort. On the other side, although we reimplemented the \textit{MIP} algorithm and our implementation might be suboptimal, \textit{MIP} requires more than $1000$ seconds to complete a run on one image.

\subsubsection{Main experiments}
\begin{table}[t]
	\newcolumntype{C}[1]{>{\centering\arraybackslash}p{#1}}
	\newcolumntype{L}[1]{>{\raggedright\arraybackslash}p{#1}}
	\newcolumntype{R}[1]{>{\raggedleft\arraybackslash}p{#1}}
	\centering
	\caption{\textbf{Main experiments.} We present the statistics of a range of different neural networks trained on different data sets. In particular, we compare the results obtained for networks trained on the \textit{plain training set} and with adversarial training (\textit{at}). Mean, minimum and maximum refer to the values of \nicefrac{$\norm{\delta_{CW}}_2$}{$\norm{\delta_{rLR-QP}}_2$} for \textit{CW} and  \nicefrac{$\norm{\delta_{DF}}_2$}{$\norm{\delta_{rLR-QP}}_2$} for \textit{DeepFool} computed on respectively the first 1000, 700 and 500 images of the test set for 1, 5 and 10 hidden layers models. The improvement rate $IR$ represents the percentage of points on which the norm of perturbation provided by our method is smaller compared to that of the other attack. The number of points where the attack, either \textit{CW} or \textit{DeepFool}, does not provide an adversarial example is indicated in brackets after the \textit{IR}. The statistics do not include these points which is an advantage for our competitors.\\}
	\begin{tabular}{C{12mm}|R{14mm} L{8mm} |C{9mm} C{9mm} C{9mm} C{11mm}|C{9mm} C{9mm} C{9mm} C{11mm}}
		\multirow{2}{*}{\centering data set}& \multicolumn{2}{c|}{\multirow{2}{*}{\centering model}} & \multicolumn{4}{c|}{\textit{CW}} & \multicolumn{4}{c}{\textit{DeepFool}}\\
		\cline{4-11}
		& & & mean & min & max & $IR$ \% & mean & min & max & $IR$ \% \\
		\hline
		\hline
		\multirow{6}{*}{\centering MNIST}& $1\times 1024$ & plain & 1.029 & 0.827 & 1.474 & 94.4 & 1.157 & 1.000 & 13.861 & $99.9(1)$\\
		& $1\times 1024$ & at & 1.066 & 0.454 & 2.434 & 83.3 & 2.107 & 1.000 & 26.179 & $100(49)$\\
		& $5\times 1024$ & plain & 1.021 & 0.496 & 1.471 & 89.3 & 1.120 & 1.001 & 6.060 & $100(8)$ \\
		& $5\times 1024$ & at & 1.013 & 0.585 & 1.880 & 75.1 & 1.404 & 1.000 & 11.009 &$100(4)$\\
		& $10\times 1024$ & plain & 0.989 & 0.704 & 1.535 & 47.8 & 1.188 & 1.000 & 1.768 & $100(5)$ \\
		& $10\times 1024$ & at & 0.997 & 0.270 & 1.507 & 67.7(61) & 2.084 & 1.000 & 5.518  &$100(4)$\\
		\ChangeRT{1pt}
		
		\multirow{6}{*}{\centering GTS}& $1\times 1024$ & plain & 1.064 & 0.936 & 2.065 & 91.0 & 1.087 & 1.000 & 15.770 & 96.6 \\
		& $1\times 1024$& at & 1.065 & 0.931 & 1.944 & 94.5 & 1.094 & 1.000 & 2.935 & 99.0 \\
		& $5\times 1024$& plain & 1.034 & 0.509 & 2.213 & 79.7 & 1.222 & 1.001 & 2.116 & 100 \\
		& $5\times 1024$& at & 1.036 & 0.314 & 3.160 & 80.1 & 1.308 & 1.000 & 3.248 & 100 \\
		& $10\times 1024$& plain & 1.023 & 0.563 & 5.067 & $76.6(1)$ & 1.346 & 1.000 & 4.254 & 99.8\\
		& $10\times 1024$& at & 1.016 & 0.533 & 2.082 & 68.4 & 1.583 & 1.003 & 3.889 & 100\\
		\ChangeRT{1pt}
		
		\multirow{6}{*}{\centering CIFAR}& $1\times 1024$ & plain & 1.090 & 0.926 & 1.802 & 98.3 & 1.056 & 1.000 & 1.966 & 99.8 \\
		& $1\times 1024$& at & 1.080 & 0.998 & 1.585 & 99.9 & 1.049 & 1.000 & 2.747 & 99.9\\
		& $5\times 1024$& plain & 1.082 & 0.825 & 1.783 & 96.0 & 1.096 & 1.001 & 4.166 & 100 \\
		& $5\times 1024$& at & 1.065 & 0.769 & 1.627 & 94.4 & 1.090 & 1.001 & 1.843 & 100 \\
		& $10\times 1024$& plain & 1.073 & 0.721 & 2.225 & 86.4 & 1.212 & 1.002 & 2.143 & 100 \\
		& $10\times 1024$& at & 1.079 & 0.644 & 1.650 & 86.4 & 1.160 & 1.000 & 1.700 & 99.8 \\
		\ChangeRT{1pt}
	\end{tabular}
	\label{tab:exps}
\end{table}
In order to evaluate our method in the task of finding adversarial examples in different settings, we choose three fully connected neural networks with increasing size and depth: $1\times 1024$ with one hidden layer of 1024 units, $5\times 1024$ having 5 hidden layers containing 1024 units each (thus, 5120 hidden units overall) and $10\times 1024$ consisting of 10 hidden layers with 1024 units each (10240 units). Every architecture is trained on three data sets: MNIST, German Traffic Sign (GTS) \cite{GTSB2012} and CIFAR-10 \cite{CIFAR10}. Moreover, for each of the previous combinations we train a model on the plain set and another integrating adversarial training (\textit{at}), using the PGD attack of \cite{MadEtAl2018}. However, since we focus on the $l_2$-norm, we adapted the implementation from \cite{Cleverhans2017} to perform the plain gradient update instead of the gradient sign (which corresponds to $l_{\infty}$-norm and thus is irrelevant for the $l_2$-norm case) on every iteration. We perform 100 iterations of the PGD attack for every batch. During the training, every batch contains 50\% of adversarial and 50\% of clean examples.

We apply both \textit{DeepFool}, in the implementation of \cite{foolbox}, and \textit{CW} $l_2$-attack. For the latter we use its untargeted formulation provided by \cite{Cleverhans2017}, keeping the settings of the original paper \cite{CarWag2016} and their code, including 20 restarts, 10000 iterations, learning rate 0.01 and initial constant of 0.001.\\
We use \textit{DeepFool} as initialization. As we noted that the solution given by \textit{DeepFool} is usually not on the surface representing the boundary between different classes, we conduct a fast binary search over the segment joining the original point $x$ and the adversarial example of \textit{DeepFool} $x+\delta_{DF}$ in order to find the decision boundary. In fact, thanks to the continuity of the classifier $f$, there exists at least one point on that segment which is still an adversarial example but closer to the clean image. Explicitly we have that \[ \exists l \quad \textrm{s.th.} \quad f_c(x)-f_l(x)>0 \quad \textrm{and}  \quad f_c(x+\delta_{DF})<f_l(x+\delta_{DF}) \] which implies \[\exists\delta^*\in \textrm{conv}(0,\delta_{DF}) \quad \textrm{s.th.} \quad f_c(x+\delta^*)=f_l(x+\delta^*).\] Then, we use the approximated $x+\delta^*$ as warm start for our method. If \textit{DeepFool} cannot provide an adversarial example, we perform the linear search between the original point and the origin of $\R^d$ to get the starting point. Moreover, for the 1 hidden layer networks we perform the search of adversarial examples over all the possible target classes, while in the case of 5 and 10 hidden layers we consider only the class the warm start belongs to.

In Table \ref{tab:exps} we report, for each setting, mean, minimum and maximum of the ratios \nicefrac{$\norm{\delta_{CW}}_2$}{$\norm{\delta_{rLR-QP}}_2$} and \nicefrac{$\norm{\delta_{DF}}_2$}{$\norm{\delta_{rLR-QP}}_2$}, where $\delta_{CW}$, $\delta_{DF}$, $\delta_{rLR-QP}$ are the perturbations provided respectively by \textit{CW} $l_2$-attack, \textit{DeepFool} and our method, and the percentage of test cases for which $\norm{\delta_{rLR-QP}}_2$ is smaller than the other attack. We consider the first 1000 images of the test sets for $1\times 1024$ models, 700 for $5\times 1024$ and 500 for $10\times 1024$.
If an attack does not find an adversarial example for a certain image we do not include it in the computations for the statistics. Notice that, unlike the other methods, \emph{rLR-QP} always provides a valid adversarial input for all test instances.

Compared to \textit{CW} attack, on 16 of the 18 models we get an average improvement between 1.3\% and 9.0\%, while for $10\times 1024$ \textit{at} on MNIST the mean is slightly below 1 but \textit{CW} attack fails to provide an adversarial example on the 12.2\% of cases. Thus it is fair to say that in only one of the 18 experiments, our method is outperformed by the \textit{CW} attack. The gap is even larger for \textit{DeepFool}, where the average gain with our method is up 110\%. Note that the benefit given by the initial binary search is not crucial to achieve this improvement, as can be seen in Table \ref{tab:iter_evol} where the statistics both before and after the linear search are shown.

\subsubsection{More iterations of \textit{rLR-QP}}
A natural question is whether allowing our method to check more linear regions can improve significantly the results or does our \emph{rLR-QP} algorithm get stuck in some suboptimal area of the input space. We consider the neural network $1\times 1024$ \textit{at} on MNIST, as it has many points where our results are worse than those of \textit{CW} attack. We want to check if for the images with $\nicefrac{\norm{\delta_{CW}}_2}{\norm{\delta_{rLR-QP}}_2}<0.95$ an improvement is possible. Then, we rerun our method, with the same hyperparameters, using the adversarial example $x+\delta_{rLR-QP}$ produced by \textit{rLR-QP} the first time as warm start for the new run. Unlike the first application, we restrict the target class to that of the starting point. Then, we iterate the same procedure a few times, always using the newest perturbation as starting point.

Table \ref{tab:iter_evol} shows the evolution of the statistics and the gradual gain in terms of quality. It is interesting that, even if the average improvement between the first and second iterations is the largest, the maximum gain is high also in the successive steps (it is of 45.7\% at the fifth application). Furthermore, the progressive increase of the mean appears almost constant and this indicates that further applications of the algorithm will lead to further improvement (for example the second largest increase of the mean of the ratios, 1.0\%, is attained the sixth time we run the algorithm). Overall, this means that running our method for a longer time (increasing $n_4$) can yield better solutions.

\begin{table}[t]
	\newcolumntype{C}[1]{>{\centering\arraybackslash}p{#1}}
	\newcolumntype{L}[1]{>{\raggedright\arraybackslash}p{#1}}
	\newcolumntype{R}[1]{>{\raggedleft\arraybackslash}p{#1}}
	\centering
	\caption{\textbf{Evolution of quality over time.} We here consider the network $1 \times 1024$ \textit{at} trained on MNIST and the 79 images which after the first application of our method have $\nicefrac{\norm{\delta_{CW}}_2}{\norm{\delta_{rLR-QP}}_2}<0.95$. We show the statistics (mean, minimum, maximum) of the ratios $\nicefrac{\norm{\delta_{CW}}_2}{\norm{\delta_{rLR-QP}}_2}$ and the number $N$ of points for which $\norm{\delta_{rLR-QP}}_2<\norm{\delta_{CW}}_2$ before performing our method (that is considering the outcome of \textit{DeepFool}, without and with the binary search) and after multiple iterations of it. Moreover, we present the progressive improvement at each step compared to the previous one. It is clear that there are improvements across all the categories and in particular for 24 of 79 points we have that the perturbation of our method at the end is smaller than the one provided by \textit{CW}.\\}
	\begin{tabular}{C{40mm}|C{12mm} C{12mm} C{12mm} C{12mm}| C{12mm} C{12mm}}
		\multirow{2}{*}{\centering step}&\multicolumn{4}{c|}{\textit{CW}}& \multicolumn{2}{c}{progr. improv.}\\
		\cline{2-7}
		& mean & min & max & $N$ & mean & max\\
		\hline
		\hline
		\textit{DeepFool} & 0.500 & $0.236^*$ & 0.893 & 0 & - & - \\
		\textit{DeepFool} + binary search & 0.511 & 0.235 & 0.893 & 0 & - & - \\
		\hdashline[0.5pt/1pt]
		I application of \textit{rLR-QP} & 0.865 & 0.454 & 0.949 & 0 & - & - \\
		II application of \textit{rLR-QP} & 0.905 & 0.454 & 1.292 & 15 & 4.9\% & 44.5\% \\
		III application of \textit{rLR-QP} & 0.908 & 0.550 & 1.292 & 16 & 0.5\% & 21.2\% \\
		IV application of \textit{rLR-QP} & 0.912 & 0.550 & 1.292 & 18 & 0.4\% & 12.9\% \\
		V application of \textit{rLR-QP} & 0.918 & 0.550 & 1.292 & 20 & 0.8\% & 45.7\% \\
		VI application of \textit{rLR-QP} & 0.927 & 0.550 & 1.292 & 23 & 1.0\% & 30.1\% \\
		VII application of \textit{rLR-QP} & 0.930 & 0.550 & 1.292 & 24 & 0.4\% & 31.4\% \\
		\hline
	\end{tabular}
	\flushleft
	$^*$ for one point \textit{DeepFool} does not produce an adversarial input, hence we perform the binary search on the segment joining the original image and the origin of $\R^d$. We do not include this point in the statistics for \textit{DeepFool} and this is why the minimum after ``\textit{DeepFool} + binary search'' is worse than after only \textit{DeepFool}.
	\label{tab:iter_evol}
\end{table}

\subsubsection{Implementation details}
Our algorithm is coded in MATLAB, relies on Gurobi solver \cite{gurobi} for the optimization processes and runs at the moment on CPUs. As it is stated in \cite{CarWag2016} a direct comparison of the runtime of different methods might be misleading as both implementation and hardware used are hardly similar. We would like to point out that our technique does not benefit at the moment of any kind of parallelization. It is however likely that major improvements in computational speed can be obtained by an adaptation that allows to exploit completely the hardware potential, including GPUs. 

\section{Outlook}
From the implementation side we will implement the current method using GPUs or if a full GPU implementation is unreasonable do a mixed CPU-GPU implementation.
The use of GPUs will then allow also to extend the method easily to convolutional neural networks and related variants which also yield continuous piecewise affine functions
if only max- and avg-pooling are used for the convolutional layers and ReLU is used for the fully connected layers.

\bibliographystyle{splncs04}

\begin{thebibliography}{10}
\providecommand{\url}[1]{\texttt{#1}}
\providecommand{\urlprefix}{URL }
\providecommand{\doi}[1]{https://doi.org/#1}

\bibitem{AroEtAl2018}
Arora, R., Basuy, A., Mianjyz, P., Mukherjee, A.: Understanding deep neural
  networks with rectified linear unit. In: ICLR (2018)

\bibitem{AthEtAl2018}
Athalye, A., Carlini, N., Wagner, D.A.: Obfuscated gradients give a false sense
  of security: Circumventing defenses to adversarial examples (2018), preprint,
  arXiv:1802.00420

\bibitem{CarEtAl2017}
Carlini, N., Katz, G., Barrett, C., Dill, D.L.: Provably minimally-distorted
  adversarial examples (2017), preprint, arXiv:1709.10207v2

\bibitem{CarWag2017}
Carlini, N., Wagner, D.: Adversarial examples are not easily detected:
  Bypassing ten detection methods. In: ACM Workshop on Artificial Intelligence
  and Security (2017)

\bibitem{CarWag2016}
Carlini, N., Wagner, D.A.: Towards evaluating the robustness of neural
  networks. In: {IEEE} Symposium on Security and Privacy. pp. 39--57 (2017)

\bibitem{GooShlSze2015}
Goodfellow, I.J., Shlens, J., Szegedy, C.: Explaining and harnessing
  adversarial examples. In: ICLR (2015)

\bibitem{gurobi}
Gurobi~Optimization, I.: Gurobi optimizer reference manual (2016),
  \url{http://www.gurobi.com}

\bibitem{HeiAnd2017}
Hein, M., Andriushchenko, M.: Formal guarantees on the robustness of a
  classifier against adversarial manipulation. In: NIPS (2017)

\bibitem{HuaEtAl2016}
Huang, R., Xu, B., Schuurmans, D., Szepesvari, C.: Learning with a strong
  adversary. In: ICLR (2016)

\bibitem{KatzEtAl2017}
Katz, G., Barrett, C., Dill, D., Julian, K., Kochenderfer, M.: Reluplex: An
  efficient smt solver for verifying deep neural networks. In: CAV (2017)

\bibitem{CIFAR10}
Krizhevsky, A., Nair, V., Hinton, G.: Cifar-10 (canadian institute for advanced
  research) \url{http://www.cs.toronto.edu/~kriz/cifar.html}

\bibitem{KurGooBen2016a}
Kurakin, A., Goodfellow, I.J., Bengio, S.: Adversarial examples in the physical
  world. In: ICLR Workshop (2017)

\bibitem{LiuEtAl2016a}
Liu, Y., Chen, X., Liu, C., Song, D.: Delving into transferable adversarial
  examples and black-box attacks. In: ICLR (2017)

\bibitem{MadEtAl2018}
Madry, A., Makelov, A., Schmidt, L., Tsipras, D., Valdu, A.: Towards deep
  learning models resistant to adversarial attacks. In: ICLR (2018)

\bibitem{MonEtAl2014}
Montufar, G., Pascanu, R., Cho, K., Bengio, Y.: On the number of linear regions
  of deep neural networks. In: NIPS (2014)

\bibitem{MooEtAl2016}
Moosavi-Dezfooli, S., Fawzi, A., Fawzi, O., Frossard, P.: Universal adversarial
  perturbations. In: CVPR (2017)

\bibitem{Cleverhans2017}
Papernot, N., Carlini, N., Goodfellow, I., Feinman, R., Faghri, F., Matyasko,
  A., Hambardzumyan, K., Juang, Y.L., Kurakin, A., Sheatsley, R., Garg, A.,
  Lin, Y.C.: cleverhans v2.0.0: an adversarial machine learning library (2017),
  preprint, arXiv:1610.00768

\bibitem{PapEtAl2016a}
Papernot, N., McDonald, P., Wu, X., Jha, S., Swami, A.: Distillation as a
  defense to adversarial perturbations against deep networks. In: IEEE
  Symposium on Security \& Privacy (2016)

\bibitem{RagSteLia2018}
Raghunathan, A., Steinhardt, J., Liang, P.: Certified defenses against
  adversarial examples. In: ICLR (2018)

\bibitem{foolbox}
Rauber, J., Brendel, W., Bethge, M.: Foolbox: A python toolbox to benchmark the
  robustness of machine learning models

\bibitem{MooFawFro2016}
S.-M. Moosavi-Dezfooli, A.~Fawzi, P.F.: Deepfool: a simple and accurate method
  to fool deep neural networks. In: CVPR. pp. 2574--2582 (2016)

\bibitem{GTSB2012}
Stallkamp, J., Schlipsing, M., Salmen, J., Igel, C.: Man vs. computer:
  Benchmarking machine learning algorithms for traffic sign recognition. Neural
  Networks  \textbf{32},  323--332 (2012)

\bibitem{SzeEtAl2014}
Szegedy, C., Zaremba, W., Sutskever, I., Bruna, J., Erhan, D., Goodfellow, I.,
  Fergus, R.: Intriguing properties of neural networks. In: ICLR. pp.
  2503--2511 (2014)

\bibitem{TjeTed2017}
Tjeng, V., Tedrake, R.: Verifying neural networks with mixed integer
  programming (2017), preprint, arXiv:1711.07356v1

\bibitem{WonKol2018}
Wong, E., Kolter, J.Z.: Provable defenses against adversarial examples via the
  convex outer adversarial polytope (2018), preprint, arXiv:1711.00851v2

\bibitem{YuanEtAL2017}
Yuan, X., He, P., Zhu, Q., Bhat, R.R., Li, X.: Adversarial examples: Attacks
  and defenses for deep learning (2017), preprint, arXiv:1712.07107

\end{thebibliography}

\end{document}